\pdfoutput=1

\documentclass[11pt]{article}

\usepackage{latex/acl}

\usepackage{times}
\usepackage{latexsym}

\usepackage[T1]{fontenc}

\usepackage[utf8]{inputenc}

\usepackage{microtype}

\usepackage{inconsolata}
\usepackage{tabularx}

\usepackage{booktabs}
\usepackage{graphicx}
\usepackage{ amssymb }

\usepackage{soul}

\usepackage{fixltx2e}

\usepackage{todonotes}

\usepackage{hyperref}

\newcounter{todocounter}

\title{QA-Noun: Representing Nominal Semantics via Natural Language Question-Answer Pairs }

\newcommand{\authorspace}{\hspace{9pt}}

\author{Maria Tseytlin$^{1,2}$ \authorspace Paul Roit$^2$ \authorspace Omri Abend$^1$ \authorspace
Ido Dagan$^2$ \authorspace Ayal Klein$^3$ \\
{$^1$Hebrew University of Jerusalem}\authorspace{$^2$Bar-Ilan University}\\{$^3$Ariel University}\\
 \footnotesize{\texttt{maria.tseytlin@mail.huji.ac.il }}
} 

\begin{document}
\maketitle

\begin{abstract}

Decomposing sentences into fine-grained meaning units is increasingly used to model semantic alignment. 
While QA-based semantic approaches have shown effectiveness for representing predicate-argument relations, they have so far left noun-centered semantics largely unaddressed.
We introduce QA-Noun, a QA-based framework for capturing noun-centered semantic relations.
QA-Noun defines nine question templates that cover both explicit syntactical and implicit contextual roles for nouns, producing interpretable QA pairs that complement verbal QA-SRL.
We release detailed guidelines, a dataset of over 2,000 annotated noun mentions, and a trained model integrated with QA-SRL to yield a unified decomposition of sentence meaning into individual, highly fine-grained, facts.
Evaluation shows that QA-Noun achieves near-complete coverage of AMR's noun arguments while surfacing additional contextually implied relations, and that combining QA-Noun with QA-SRL yields over 130\% higher granularity than recent fact-based decomposition methods such as FactScore and DecompScore.
QA-Noun thus complements the broader QA-based semantic framework, forming a comprehensive and scalable approach to fine-grained semantic decomposition for cross-text alignment.
\end{abstract}

\section{Introduction}
\label{sec:intro}


\begin{figure}[t]
    \resizebox{\columnwidth}{!}{%
    \includegraphics[width=\columnwidth]{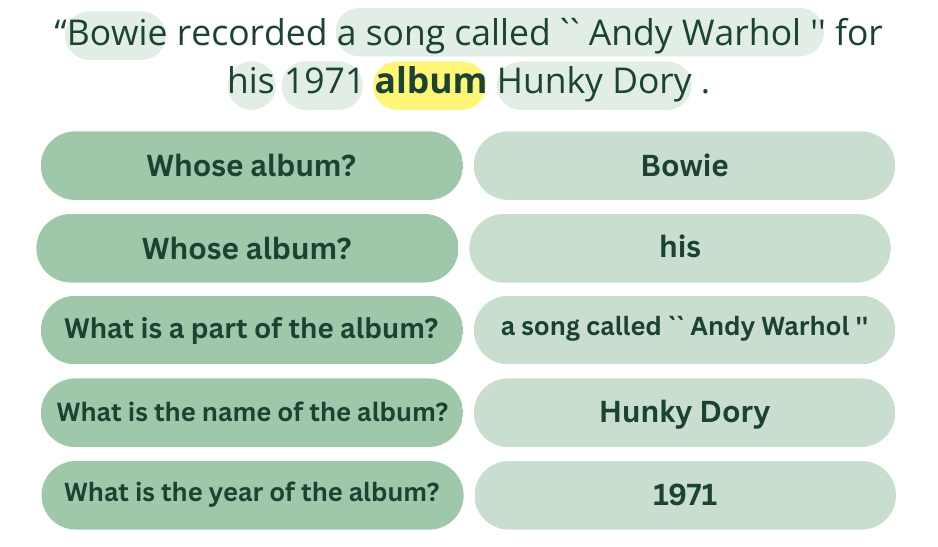}
    }
    \centering
    \caption{Example QA-Noun annotations for a single target noun (\textbf{\hl{highlighted}}), explicating  each ``atomic'' fact involving the noun as an individual QA pair.}
    \label{fig:leading_example}
\end{figure}

Semantic representations of text, which decompose sentence meaning into smaller information units, are increasingly recognized as essential for modeling fine-grained alignment between texts. 
Such decomposition based alignments are beneficial for assessing the faithfulness of generated texts against a source of truth, highlighting which parts were preserved, omitted or hallucinated \cite{fan-etal-2023-evaluating,qiu-etal-2024-amrfact,chen2023propsegmentlargescalecorpuspropositionlevel}, and they can also be used to guide the generative process with better content selection \cite{narayan2023conditionalgenerationquestionansweringblueprint,Zhang2025QAPyramid}. 

To meet these needs, recent work has explored decomposing text into discrete meaning units, either in interrogative form \citep{deutsch-etal-2021-qas4eval_summ,durmus-etal-2020-feqa,honovich2021qgqa4factuality_eval} or as declarative “atomic facts” \citep{min-etal-2023-factscore,wanner-etal-2024-decompscore}.
These approaches, while effective for downstream evaluation, lack an underlying representation framework for systematically covering all fine-grained information units. 
To overcome these issues, some recent work has revisited traditional NLP semantic formalisms, such as Semantic Role Labeling (SRL) and Abstract Meaning Representation, which rely on formal semantic schemata to model meaning via labeled predicate–argument relations 
\citep{qiu-etal-2024-amrfact,fan-etal-2023-evaluating}.

An alternative semantic representation paradigm, based on natural-language question–answer (QA) pairs, was pioneered by \textbf{QA-SRL} \citep{he-etal-2015-question}. Instead of relying on formal role inventories, QA-SRL expresses predicate–argument relations through simple, intuitive questions (e.g., “Who did X?”) answered with spans from the sentence. This format is easily interpretable for lay annotators, supports efficient crowdsourcing, scales across domains and languages, and aligns naturally with language models --- making it well-suited for both human and automated model-driven semantic annotation.
Recently, this method was expanded to target deverbal nominalizations \cite{klein-etal-2020-qanom}, discourse relations \cite{pyatkin-etal-2020-qadiscourse} and adjectives \cite{pesahov-etal-2023-qa}.
However, nominal relations expressed between a noun and its arguments have largely been overlooked.

This work introduces \textbf{QA-Noun}, a QA-based representation for nominal semantics that is both layman-attainable and semantically comprehensive --- capturing not only explicit grammatical roles, but also implicit relations inferred from context.
QA-Noun targets nine core semantic dimensions relevant to nouns, each instantiated through a corresponding question template (See Figure~\ref{fig:leading_example} for an illustrative example). 
This design enables interpretable, highly fine-grained decomposition into minimal facts involving the noun --- where each QA pair corresponds to a single semantic relation. 
Notably, decomposition granularity is crucial for accurately modeling semantic alignment across texts since misalignments can occur for any minimal individual fact. 
For example, in Figure~\ref{fig:leading_example}, each QA pair represents an atomic fact whose faithfulness to a source must be assessed independently.

We make the following contributions:
(1) we extend the QA-based semantic paradigm to cover noun semantics, complementing verbal QA-SRL to enable an exhaustive, fine-grained decomposition of sentence meaning;  
(2) we design a novel annotation framework for nouns, capturing nine core semantic dimensions through  interpretable question templates (\S \ref{sec:method});  
(3) we release a high-quality dataset of noun-centered QAs (\S \ref{sec:dataset_construction}) and assess its consistency and coverage (\S \ref{sec:data_quality_assessment});  
(4) we develop and evaluate QA-Noun models and integrate them with QA-SRL into a unified decomposition tool (\S \ref{sec:model}); and  
(5) we show that combining QA-Noun with QA-SRL yields  a 130\%--150\% gain in semantic granularity over prior fact-based decomposition methods (\S \ref{sec:granularity_assessment}).

\section{Background}
\label{sec:background}

\subsection{Semantic Representations of Nouns}
Different semantic formalisms such as PropBank \cite{palmer-etal-2005-proposition} and NomBank \cite{meyers-etal-2004-annotating} have sought to capture some of the information that a text conveys within a structured scheme. 
NomBank annotates nominal predicates in the Penn Treebank with their arguments, assigning each a semantic role from a fixed inventory. 
For example, in the phrase \textit{``higher \textbf{rate} of improvement''}, NomBank labels  \textit{improvement} as the \texttt{theme}, and \textit{higher} as the \texttt{value} of the predicate \textbf{rate}.
We use NomBank as a key reference in the design of QA-Noun: its argument structures informed our initial question template set and served as a basis for assessing coverage during the design phase.

A later and widely adopted framework is \textbf{Abstract Meaning Representation} \citep[AMR;][]{banarescu-etal-2013-abstract}, which encodes sentence-level semantics as rooted, labeled graphs. AMR extends beyond predicate–argument structures to capture a broader range of semantic relations, including nominal ones. 
Its role inventory combines the PropBank roleset for eventive predicates with general relations such as \texttt{:poss}, \texttt{:part}, and \texttt{:consist-of}, enabling the representation of non-eventive and relational nouns. Because of its broad scope and community adoption, AMR is a central comparison point for QA-Noun. 
In Section~\ref{subsec:comparisons}, we show that our question templates capture essentially all noun arguments annotated in AMR, and often surface additional, contextually implied relations.

NomBank and AMR represent key prior efforts in modeling nominal semantics but rely on formal role inventories and expert annotation, which limits scalability.
\textbf{NounAtlas} \citep{navigli2024nounatlas} expands nominal SRL coverage through large-scale automatic projection and clustering of argument structures within the traditional SRL paradigm.
\textbf{TNE} \citep{elazar-etal-2022-text} instead leverages natural-language prepositions as intuitive relation labels to annotate noun phrase relations at scale.
Neither, however, is designed to provide a predicate–argument representation of nouns in context as QA-Noun does.

\subsection{Semantic QA Approach}

To avoid formal, hard-to-scale role inventories, QA-SRL \cite{he-etal-2015-question} introduced a natural language–based representation in which arguments and their roles are expressed as question–answer pairs. In this formulation, the question encodes the semantic role, while the answers identify the corresponding arguments. This format does not depend on predefined semantic role lexicons, can be explained to annotators with minimal training due to its use of natural language, and captures valuable implicit arguments that may not be explicit in syntax \citep{roit-etal-2020-controlled}.

Building on QA-SRL, the paradigm has gradually expanded into a broader \textbf{QA-based semantics (QASem)} framework. This includes extensions to deverbal nominalizations \citep{klein-etal-2020-qanom}, adjectives \citep{pesahov-etal-2023-qa}, and discourse relations \citep{pyatkin-etal-2020-qadiscourse}, moving toward a unified question–answer representation for predicate–argument structure.

Beyond annotation, QASem has proven effective as a semantic decomposition layer for downstream tasks. Recent work has leveraged QA-based predicate–argument units for fine-grained cross-text alignment and evaluation:
QAAlign \citep{brook-weiss-etal-2021-qa} aligns information across texts via QA pairs; 
\citet{roit-etal-2024-explicating} leverage the induced QA-SRL grammar to detect arguments across sentences;
QAPyramid \citep{Zhang2025QAPyramid} uses QASRL/QANom units for pyramid-style content selection evaluation in summarization; and
\citet{cattan2024localizingfactualinconsistenciesattributable} apply QA-SRL and QANom to localize factual inconsistencies in attributable text generation.
Together, these studies demonstrate that QA-based predicate–argument representations provide an intuitive and fine-grained decomposition of meaning that supports evaluating faithfulness, information selection, and source attribution.

In this work, we extend this QA-based paradigm to cover a wide range of noun-centered semantic relations, complementing QA-SRL’s verbal focus and completing a major step toward comprehensive QA-based semantic decomposition.

\subsection{Granular Semantic Decomposition}
\label{subsec:bg_decomp}
A growing body of work in text generation and evaluation has highlighted the benefits of decomposing sentences into smaller, interpretable meaning units to enable precise cross-text alignment.
The \textbf{QG-QA} framework exemplifies this trend by representing sentences as sets of question–answer pairs, supporting fine-grained evaluation of semantic overlap and faithfulness in summarization and factuality tasks \citep{eyal-etal-2019-qa4eval_summ, gavenavicius2020qas4eval_summ, deutsch-etal-2021-qas4eval_summ, honovich2021qgqa4factuality_eval, durmus-etal-2020-feqa}.

In a similar spirit, recent systems such as \textbf{FactScore} \citep{min-etal-2023-factscore}, \citep{zhu2024reducinghallucinations} and \textbf{DecompScore} \citep{wanner-etal-2024-decompscore} decompose sentences into sets of “atomic” natural-language facts for factuality and content selection evaluation. These approaches demonstrate that finer-grained decompositions yield more accurate and interpretable measures of semantic consistency. However, they typically rely on prompting large language models without a defined schema, which makes the granularity and coverage of the resulting “atomic facts” difficult to control in a principled way.

Among these, DecompScore goes further by formalizing the decomposition task and introducing a metric for \textit{atomicity}, or granularity, to compare decomposition methods. The measure counts the number of units that can be faithfully inferred from the source text, thereby favoring approaches that produce a larger set of accurate, fine-grained meaning units. 
\textbf{CORE} \citep{jiang-etal-2024-core} extends this by postprocessing decompositions to remove cross-unit redundancies, ensuring that the metric is not inflated by overlapping or paraphrastic facts.
In this paper, we adopt the DecompScore granularity measure, combined with CORE’s redundancy control, to assess QA-Noun together with QA-SRL, showing substantial gains in granularity compared to prior fact-based decomposition methods (\S\ref{sec:granularity_assessment}).

While recent work advocates decomposing text into fact-like units, classical linguistically oriented representations --- such as SRL, dependency-based semantics, AMR, and others --- model textual meaning through structured predicate–argument relations \citep{abend2017state}. These frameworks are grounded in \textbf{Neo-Davidsonian semantics} \citep{parsons1990events}, where events and entities are represented as variables linked by binary relations labeled with thematic roles. 
For example, “The president signed the bill” is represented with an event \texttt{sign(e)}, entities \texttt{president(p)} and \texttt{bill(b)}, and relations \texttt{Agent(e,p)} and \texttt{Theme(e,b)}. 
QA-Noun builds on this tradition in a natural-language QA format, providing question–answer representations for nominal relations that, together with verbal QA-SRL, yield a structured, fine-grained decomposition of sentence meaning grounded in predicate–argument structure.

\section{The QA-Noun Task}
\label{sec:method}

\begin{figure}[t]
    \includegraphics[width=0.5\textwidth]{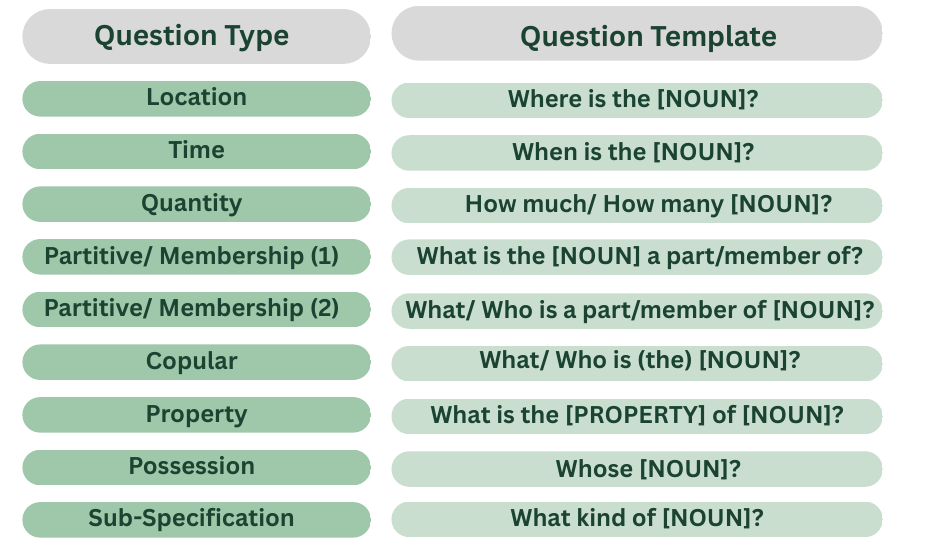}
    \caption{Example questions illustrating QA-Noun question templates. NOUN refers to the target noun.}
    \label{fig:question_templates}
\end{figure}
We introduce the QA-Noun task: a structured approach for identifying the semantic arguments of nouns in context. 
Given a sentence and a marked noun, our goal is to identify other phrases from the sentence that pertain to the noun, and  represent its semantic relation to the noun using a simple question. 
For example, see \autoref{fig:leading_example} for the arguments and their corresponding roles for the noun \textbf{album}.

In our task, we represent common semantic relations expressed by nouns through questions generated from a carefully designed set of templates.

To ensure broad applicability, we place no restrictions on the types of nouns considered as predicates: any noun, regardless of its lexical category, or role in the sentence, is treated as a potential noun predicate. 

\subsection{Question Templates.}

We define nine core question templates to capture the main argument types expressed by nouns (\autoref{fig:question_templates}).
Examples include:

\begin{tabular}{ll}
\texttt{Possessive} -- & his \textit{wife’s} late \textbf{aunt}. \\
\texttt{Locative} --   & the \textit{Paris} \textbf{bridge}.     \\
\texttt{Partitive} -- & an \textit{army} \textbf{officer}.     
\end{tabular}

Our design follows the QASem tradition of systematically crafted question templates to capture predicate–argument relations \citep{he-etal-2015-question, pyatkin-etal-2020-qadiscourse, klein-etal-2020-qanom, pesahov-etal-2023-qa}.
The resulting templates provide interpretable and controllable mappings between linguistic structure and semantic role expression, making them well-suited for both annotation and model supervision.

\paragraph{Flexible Template Realization.}
While most templates are fixed, several allow minor modifications to better fit the sentence context and improve naturalness.
For example, the \texttt{Partitive} template \textit{\textbf{What/Who is a part/member of [NOUN]?}} permits annotators to choose between \textit{part} or \textit{member}, and between \textit{what} or \textit{who}, depending on the noun’s semantics and discourse context.
This controlled flexibility preserves the discrete, role-oriented character of the framework while enabling smoother phrasing and context-sensitive adaptation.

\paragraph{Hybrid Labeling via the Property Template.}
One template, \texttt{Property}, is reserved for open-ended attributes.
It introduces a placeholder for a context-specific descriptor drawn from an open vocabulary, enabling flexible coverage of semantic properties beyond fixed roles.
For example, in the sentence \textit{``Valley Ranch is the team's 30-acre practice \textbf{camp}''}, the property template produces: \\
\textit{What is the [purpose] of the camp?} →  \textit{practice} \\
\textit{What is the [size] of the camp?} →  \textit{30-acre} \\ 
Typical property values include \textit{name}, \textit{purpose}, \textit{cause}, and \textit{status}, inferred directly from context.

This hybrid design uniquely combines discrete, interpretable question types with an open-ended semantic slot.
The result is a labeling scheme that offers both \textbf{consistency} (through fixed question types) and \textbf{expressivity} (via open-vocabulary properties), bridging the gap between structured role labeling and fully free-form natural language semantics \citep{michael2018qamr}.

\paragraph{Template Development and Validation.}
To ensure comprehensive coverage of noun-centered semantic relations, we drew on prior linguistic resources on nouns and noun compounds \citep{meyers-etal-2004-annotating, tratz-hovy-2010-taxonomy}.
Starting from NomBank-aligned categories, we iteratively refined the templates through controlled annotation rounds.
In early stages, we experimented with a larger set of candidate templates and crowdsourced dozens of sentences with multiple workers, analyzing the resulting \textbf{confusion matrix} of annotator choices to identify overlapping or ambiguous roles (e.g., between \texttt{Partitive} and \texttt{Membership}).
We then abstracted and merged such categories to achieve a more discrete and distinguishable inventory, repeating this process until the final nine-template set reached stable coverage and low annotator confusion.

\subsection{Argument and Question Scope} 
In QA-Noun, each semantic argument is represented as a contiguous phrase, that answers one of our template-based questions. 
The QA-Noun task is designed to complement verbal SRL by focusing specifically on noun-centered semantic relations that are not addressed through verb-based annotation. 
Thus we refrain from annotating arguments that would otherwise have been included in a semantic analysis of verbs in the sentence.

Annotators are instructed to select the most specific question template appropriate to the context, ensuring that the assigned role precisely captures the semantic relation of the argument to the noun. 
Nevertheless, multiple questions can often validly apply to the same argument (e.g., \textit{What is the location of X?} vs.\ \textit{Where is X?}). 
We view this as an inherent feature of using natural language to represent semantics rather than a deficiency: each overlapping QA provides a complementary perspective on the relation. 
Downstream systems may aggregate such overlapping labels or exploit them as multi-faceted evidence for richer semantic modeling.

\section{Dataset Construction}
\label{sec:dataset_construction}

\paragraph{Data} To create the QA-Noun dataset, we annotated over 1600 sentences with over 2000 noun mentions\footnote{In each sentence, we identify the target nouns using SpaCy’s POS-tagger.} across two main domains: Wikinews and Wikipedia. 
We annotated 1,686 sentences encompassing 2,029 nominal predicates, yielding a total of 4,869 arguments. 
The dataset was split into 50/10/40 (\%) for the train, development and test splits.
See \autoref{tab:templates_stat} for template statistics.

\begin{table*}[t!]
\resizebox{\textwidth}{!}{%
\begin{tabular}{llllllllllll}
\hline
      & Property    & Possession   & Location   & Quantity   & Partitive/ Membership (1)   & Partitive/ Membership (1)   & Copular   & Sub-Specification   & Time   \\ \hline
Total & 1146 & 740 & 290 & 184 & 586 & 600 & 302 & 921 & 140 \\ \hline
\end{tabular}%
}
\caption{Template statistics of the QA-Noun dataset}
\label{tab:templates_stat}
\end{table*}

\begin{figure}[t!]
    \centering
    \resizebox{\columnwidth}{!}{%
    \includegraphics[width=\columnwidth]{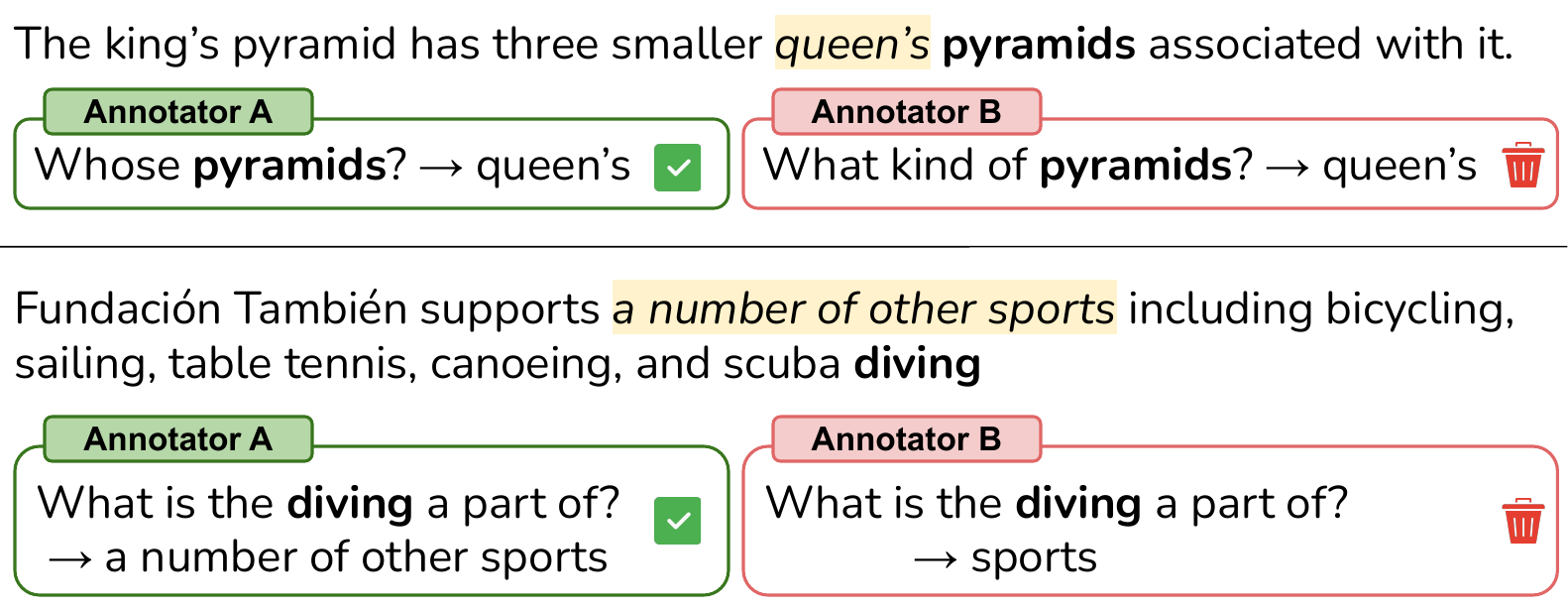}
    }
    \caption{Reconciliation phase between two QA-Noun annotators. The annotators adjudicate between two proposed arguments that disagree either by extent or by semantic role. The selected argument and role after adjudication is shown schematically under annotator A (in green), while the discarded argument-role is shown under annotator B (in red). The top example showcases role (question) disagreement between the two annotators, while the bottom example depicts different extents (phrases) of the same argument.}
    \label{fig:reconciliation} 
\end{figure}

\paragraph{Annotation Process} 
We employed in-house annotators, primarily linguistics students or experienced English users (e.g., writers and language instructors). 
Following a controlled onboarding procedure inspired by \citet{roit-etal-2020-controlled}, candidates first underwent screening for English proficiency and fluency, then completed a paid training phase. 
During training, annotators studied detailed task guidelines\footnote{The QANoun guidelines are publicly available at \href{https://docs.google.com/presentation/d/18rrWQNg6lrOawSvDVptDlDs4k3g39G_gE0tV-EQfPBE/edit?usp=sharing}{this slideshow}.}, reviewed illustrative examples, and annotated a practice set of several dozen examples drawn from an \textit{expert set} of 80 gold instances. 
Each candidate received personalized feedback based on automatic comparisons to gold annotations and follow-up meetings with the first author. 
This process ensured consistent, high-quality annotation before contributing to the main dataset.

During pilot studies, we observed, similarly to \citet{roit-etal-2020-controlled}, that argument identification from individual crowd workers often lacked sufficient coverage.
To address this problem, we employ a two-annotator protocol to annotate a single predicate.
First, two trained annotators are given a shared set of target predicates.
Each annotator \textit{independently} produces QA-pairs for each noun in the set according to our guidelines, using a dedicated annotation interface designed to streamline question formulation and answer span selection (see Appendix \ref{sec:appendix-annotation-interface} for details).

\paragraph{Consolidation} After independent annotation, each pair of annotators meets online to reconcile differences.
They review each other’s argument spans and question templates, resolving discrepancies to reach a single agreed set.
Missed arguments can be added, erroneous spans discarded, and questions refined to better capture the noun–argument relation.
The final QA pairs are thus double-verified for accuracy while combining the coverage of two independent passes.
\autoref{fig:reconciliation} illustrates this process: in one case, the more context-specific question is selected; in another, the more precise span is retained.

  

We employ this two-step protocol to collect our high-quality evaluation benchmark, while our training data is collected using a single trained annotator. This follows common practice in semantic annotation \citep[e.g.,][]{roit-etal-2020-controlled, fitzgerald-etal-2018-large, kwiatkowski2019natural}, where the training data undergoes lighter quality control to enable greater diversity and scale, while the evaluation sets are double-annotated and adjudicated to ensure reliability. Details regarding annotator compensation and cost breakdown are provided in Appendix~\ref{sec:appendix-annotation-costs}.

\section{Assessing QA-Noun Dataset Quality}
\label{sec:data_quality_assessment}

\subsection{Evaluation Metrics}
\label{subsec:eval_metrics}
\begin{figure*}[t!]
    \centering
    \resizebox{\textwidth}{!}{%
    \includegraphics[width=\textwidth]{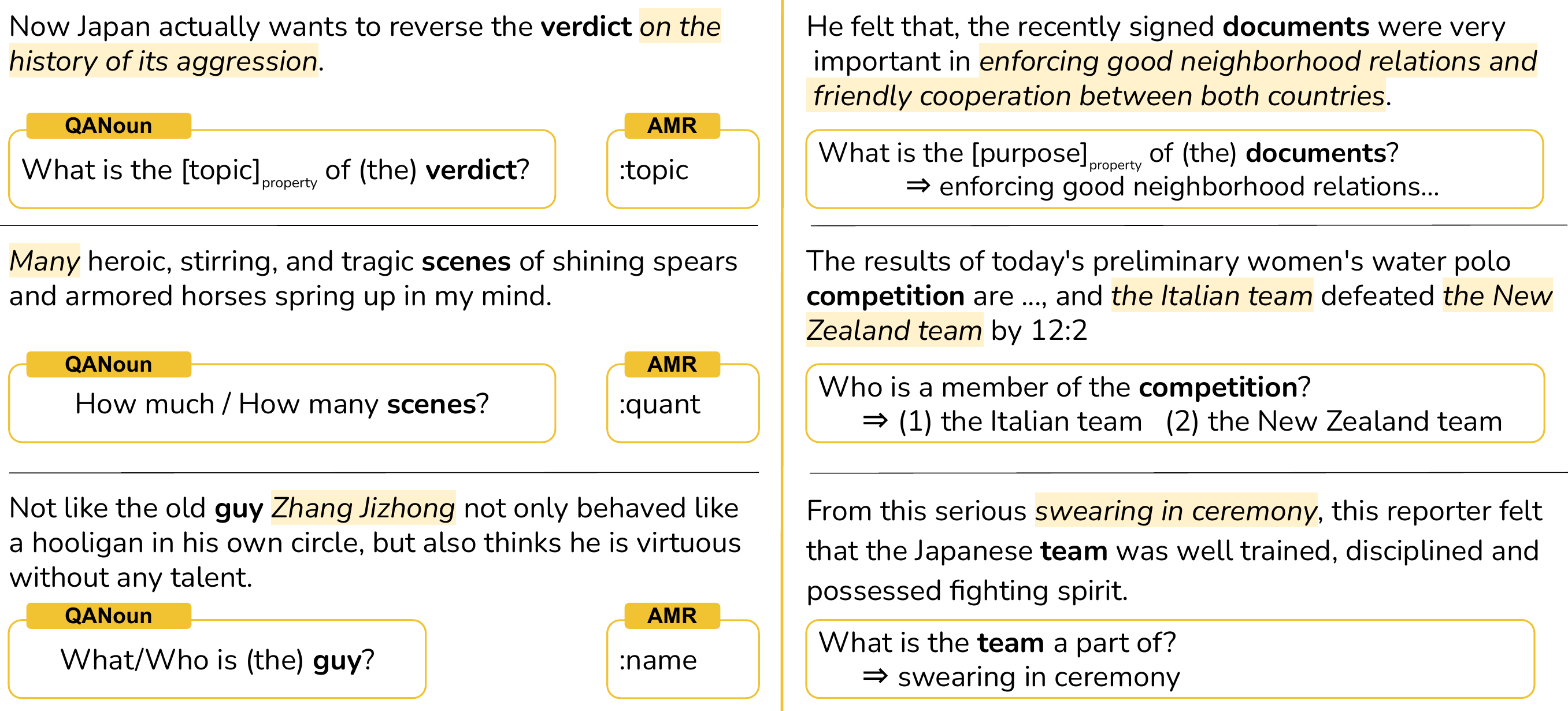}
    }
    \caption{Comparison between example sentences with AMR and QA-Noun annotations. The noun predicate in each sentence is marked in bold and its argument is highlighted inline. \textbf{Left} Comparison between semantic roles when the argument is mutually annotated. \textbf{Right} Diverse arguments captured by QA-Noun's annotators that were out of scope for AMR. They represent different implied meanings, memberships and other relations.}
    \label{fig:amr_comparison} 
\end{figure*}
Similar to SRL, QA-Noun evaluation measures two abilities --- correctly detecting the noun's arguments, and the correct assignment of semantic roles to these arguments.
Following previous work \cite{pesahov-etal-2023-qa,roit-etal-2020-controlled,pyatkin-etal-2020-qadiscourse,klein-etal-2020-qanom}, we report standard precision and recall scores for unlabeled argument detection (UA) against a set of ground-truth arguments.
Briefly, a predicted argument is considered to be correct if it significantly overlaps with a gold argument, with a token-level intersection over union greater than 0.5.
We apply maximal bipartite matching to enforce a one-to-one alignment between predicted and gold arguments, weighting each pair by their overlap score.
We count the number of true positives as the number of matches, and false positives and false negatives as the number of leftover arguments from the predicted and ground truth sets, respectively.

In contrast to argument detection, evaluating the accuracy of semantic role assignment is a challenge. 
In QA-Noun, the roles are represented as question templates, and they are not mutually exclusive.
For instance, as shown in \autoref{fig:leading_example}, the argument \textit{1971} could be annotated both with \textit{What is the [year] of the album?} and \textit{When is the album?}.
Therefore, comparing against a single ground-truth template could underestimate role assignment accuracy. 

To address this, we manually evaluate whether the selected question template accurately captures the semantic relationship between the noun and the predicted answer span.
This evaluation is performed on correctly predicted arguments, and we report the proportion of \textit{sound role assignments} (SRA).
To further increase reliability, two experts independently assess each role assignment, and report the average SRA from their evaluations.  

\subsection{Comparison with AMR}
\label{subsec:comparisons}

We compared QA-Noun annotations and AMR structures over the same set of nouns to gain insight about their relative coverage in the scope of annotation.
Given a predicate noun, we manually align its QA-Noun arguments with the associated AMR entities  and analyze the differences.
Since AMR represents the sentence using a directed graph over \textit{entities}, while QA-Noun annotates \textit{lexical units} in the sentence, we first identify the node in the AMR graph that represents the predicate entity and extract its arguments.\footnote{At times, the noun's direct parent in AMR refers to the same entity, and in that case we take the parent's arguments as well. }
In particular, we consider both directions to and from the predicate node in the AMR graph as plausible arguments.

In this analysis, we annotated a sample of 40 nouns from the AMR Bank, yielding 156 QA-Noun arguments, while the corresponding AMR entities include 90 arguments in total. 
Our analysis showed that QA-Noun captures almost all noun-related relations represented in AMR (\textbf{89/90; recall = 0.99, 95\% CI [0.97, 1.00]}, bootstrap 200K replicates), with most question templates aligning closely to AMR roles (\autoref{fig:amr_comparison}).
Beyond this significant overlap, QA-Noun has annotated 65 additional arguments absent from AMR, including implied relations (21), membership roles (16), coreferent mentions (13) and various other cases, alongside 4 annotation errors. 
While some of the implied relations are inferential and out of scope for AMR, some are captured implicitly in AMR's deeper graph structure but are not linked as direct arguments. 

These promising results, both in correctness of the QA-Noun arguments and the almost full coverage of AMR entities, suggest that QA-Noun contributes reliably annotated semantic relations that AMR often leaves implicit or underspecified.
QA-Noun thus provides a broader and more accessible representation of noun arguments, including co-referential mentions and implied roles that are difficult to recover from AMR alone.
Its natural-language question format enables detailed semantic coverage while remaining intuitive and scalable for annotation, making it a strong complement to existing structured frameworks akin to AMR.

\subsection{Inter-Annotator Agreement}
\label{subsec:IAA}
To estimate the consistency of the dataset across different annotations, we measure inter-annotator agreement (IAA) on a sample of 90 target nouns. While worker-vs-worker agreement for QAs is somewhat partial --- mostly due to insufficient coverage, as discussed above (\S \ref{sec:dataset_construction}) --- the overall consistency of the dataset is assessed by comparing consolidated annotations obtained from disjoint pairs of workers after adjudication.
The macro-averaged unlabeled agreement (UA) F1 score for inter-annotator agreement is \textbf{72.8}.

Although somewhat lower than the expert agreement levels typically reported for tightly constrained schemes such as NomBank, our IAA reflects the open-ended nature of QA-Noun’s semantic coverage and the expressivity of the QA format, and is comparable to agreement levels reported for other QASem tasks \citep{klein-etal-2020-qanom, pesahov-etal-2023-qa}.

\section{Modeling}
\label{sec:model}
\begin{table}[t!]
\resizebox{\columnwidth}{!}{%
\begin{tabular}{cllll}
\toprule
     & Model                     & Precision & Recall & F1 \\ \midrule

  & Llama 3 \hfill 8B                & 56.4            & 35.5               & 43.6        \\
  \textbf{ICL}    & Llama 3 \hfill 70B               & 67.4            & 40.2               & 50.4        \\
 & Llama 3.1 \hfill 405B            & \textbf{64.7}            & 51.0               & \textbf{57.0}        \\ \midrule
           & Llama 3 \hfill 8B                & 49.7            & \textbf{62.7}               & 55.4        \\
\textbf{FT}  & Qwen 2.5 \hfill 14B              & 62.5            & 48.1               & 54.4        \\
           & Phi 4 \hfill 14B                 & 49.1            & 57.5                & 53.0        \\ 
\bottomrule
\end{tabular}
}
\caption{Different models automatic evaluation results against our test set. All reported metrics are for unlabeled argument detection (UA).  ICL stands for In-Context Learning methods, while FT is for fine-tuned methods. }
\label{tab:models_result}
\end{table}

In addition to defining the QA-Noun representation and dataset, our goal is to develop and release a practical tool for semantic decomposition --- one that is both accurate and efficient. 
To this end, we experiment with two modeling approaches: in-context prompting and parameter-efficient fine-tuning, evaluating their performance on the QA-Noun task. 
Data, models and experiments code can be found in the project repository.\footnote{\url{https://github.com/unimaria/QA-Noun}} 

\subsection{Methods}
\label{subsec:methods}

\paragraph{In-Context Learning (ICL).}  
We evaluate several large language models (LLMs) using few-shot prompting. 
Each model is prompted to generate all relevant QA pairs for a target noun in context using our predefined question templates, with at least two examples per template included in the prompt to guide completions.
The full prompt is provided in Appendix \ref{sec:appendix-model}. 
We test the following LLMs: \texttt{LLaMA-3-70B}, \texttt{LLaMA-3-8B} and \texttt{LLaMA-3.1-405B}.

\paragraph{LoRA Fine-Tuning (FT).}  
To adapt moderately sized models for the task, we apply Low-Rank Adaptation (LoRA; \citealp{hu-etal-2021-lora}), updating only a small subset of parameters during training. 
We fine-tune three models --- \texttt{LLaMA-3-8B}, \texttt{Qwen-2.5-14B}, and \texttt{Phi-4-14B} --- on our QA-Noun training set, using gold question-answer pairs as supervision. Multiple hyperparameter configurations were explored to optimize performance (see Appendix \ref{sec:appendix-ft} for details).

\subsection{Model Evaluation}
\label{subsec:model_eval}

\paragraph{Main Results.}  
\autoref{tab:models_result} presents evaluation results on our test set, comparing in-context and fine-tuned models in terms of unlabeled argument detection.
The best overall performance is achieved by \texttt{LLaMA-3.1-405B} in the in-context setting, demonstrating strong generalization even without task-specific training.
Notably, the fine-tuned \texttt{LLaMA-3-8B} performs competitively, outperforming all other fine-tuned and in-context models apart from \texttt{LLaMA-3.1-405B}.


These results highlight the value of task-specific supervision: with a modest parameter footprint, fine-tuned models approach the performance of much larger LLMs --- making them practical for large-scale decomposition pipelines where efficiency is essential. For this reason, we select the fine-tuned \texttt{LLaMA-3-8B} as our parser backbone, balancing strong performance with open licensing and cost-efficiency for scalable downstream use.

\paragraph{Role Assignment}
As discussed in Section~\ref{subsec:eval_metrics}, the flexible nature of QA-Noun question templates poses a challenge for automatic evaluation against a ground truth that contains only a single label. 
To address this, we conducted both manual and automatic evaluations of role assignment for our selected model, a fine-tuned \texttt{LLaMA-3-8B}. 

In the manual evaluation, two of the authors independently reviewed model-generated questions for correctly identified arguments across 54 target nouns (115 QA pairs in total). 
The resulting average semantic-role accuracy (SRA) was \textbf{58.5\%}, indicating that while arguments are generally recovered reliably, the model often struggles to select the most contextually appropriate question template.

To complement this small-scale analysis, we performed an automatic evaluation using a strong LLM (\texttt{GPT-4o}) as an entailment-based judge. 
For 1,425 QA pairs where the predicted argument span was correct, the model was asked whether each QA pair was entailed by the original sentence --- interpreted as a proxy for question validity. 
Approximately \textbf{65\%} of the QAs were judged valid, a slightly higher rate than the manual estimate. 
Together, these analyses suggest that while QA-Noun effectively captures most nominal relations, selecting the most fine-grained and semantically precise role remains a key challenge, motivating future work on improved modeling and richer supervision.

\paragraph{}
We next move from argument-level accuracy to evaluating QA-Noun+QA-SRL as a decomposition framework, comparing its granularity to fact-based methods such as FactScore and DecompScore.

\section{Granular Information Decomposition}
\label{sec:granularity_assessment}

\begin{figure}[t!]
    \centering
    \resizebox{\columnwidth}{!}{%
    \includegraphics[width=\columnwidth]{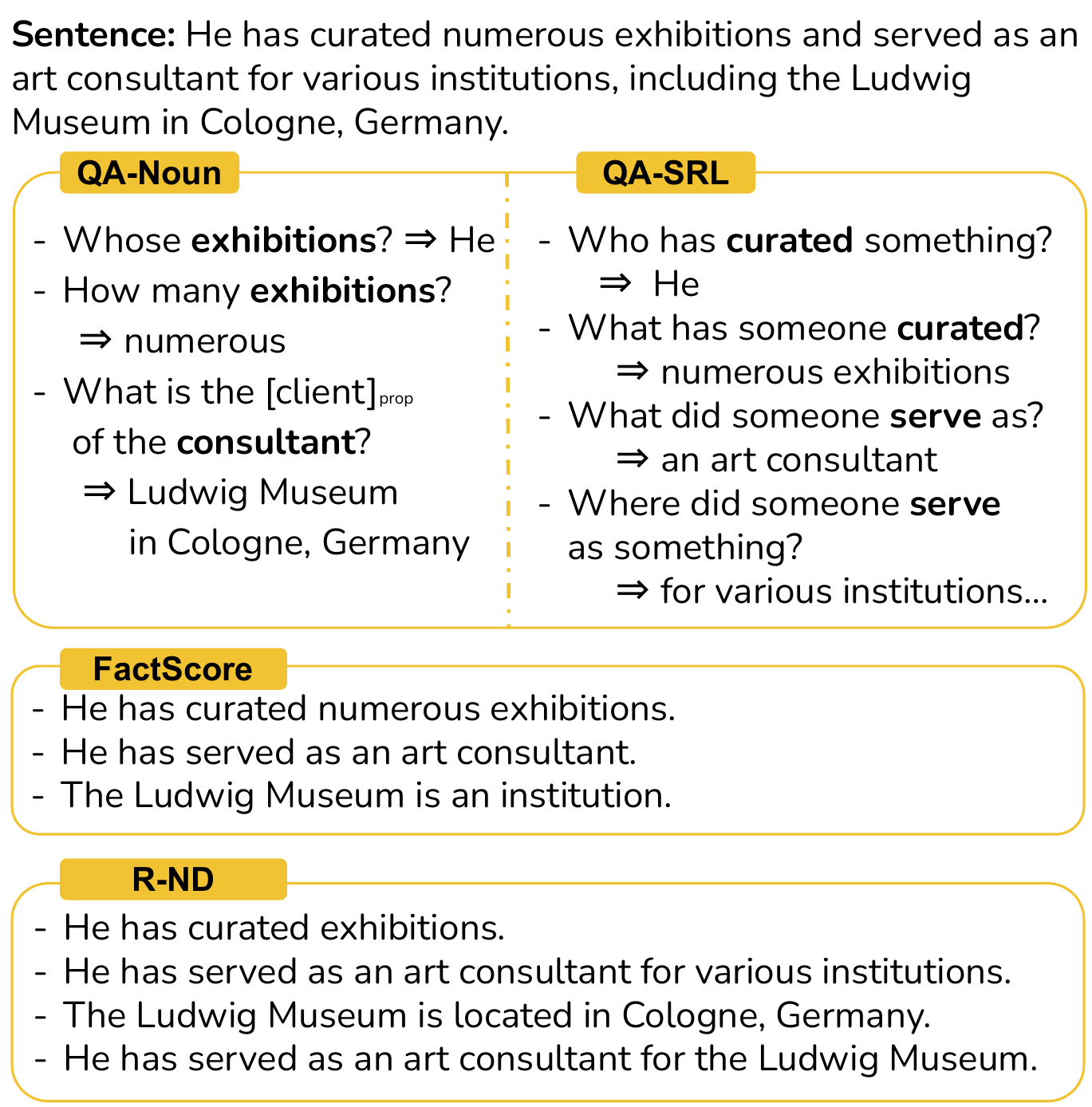}
    }
    \caption{Sentence decomposition with QA-Noun and QA-SRL compared to fact-based approaches. QA-Noun captures noun-centered relations (e.g., \textit{Whose exhibitions?}) and surfaces implicit links such as client–consultant relations, which fall under inferential arguments. Combined with QA-SRL verbal roles, they yield a structured predicate–argument breakdown. FactScore and R-ND generate declarative “atomic facts” but typically do not capture such inferential relations.}
    \label{fig:decomp_example} 
\end{figure}
\begin{table}[]
\resizebox{\columnwidth}{!}{%
\begin{tabular}{@{}lccc@{}}
\toprule
\multicolumn{4}{c}{\textbf{DecompScore}: Entailed Sub-claims Per Sentence} \\ \midrule
Method & Generated & Non-Redundant & Entailed \\
FactScore {\tiny \texttt{(GPT-4o)}} & 4.9 & 3.2 & $3.1\pm0.1$ \\
R-ND {\tiny \texttt{(GPT-4o)}} & 5.4 & 3.7 & $3.7\pm0.1$ \\
QASem {\tiny \texttt{(Llama-3-8B)}} & 14.1 & 7.2 & $\mathbf{4.8}\pm0.2$ \\ \bottomrule
\end{tabular}%
}
\caption{DecompScore's decomposition \textit{atomicity} metric: The number of sub-claims/QAs generated by each method per sentence, broken down by the number of generated, non-redundant, and finally entailed units. The \textit{Entailed} column is the final DecompScore metric, calculated with a 95\% confidence interval. Our approach (QA-Noun + QA-SRL) is denoted as QASem.}
\label{tab:decomp_score}
\end{table}


As discussed in the introduction, capturing sentence meaning via a maximally atomic decomposition of information units is key for modeling semantic alignment.
QA-Noun combined with QA-SRL produces explicit predicate–argument QA pairs as granular meaning units, in contrast to recent “atomic fact” approaches that prompt LLMs to generate unconstrained declarative statements.
To quantify this difference, we evaluate the granularity of our QA-based decompositions using \textbf{DecompScore} \citep{wanner-etal-2024-decompscore}, a metric designed to assess the granularity and coverage of decomposition methods, as described in Section \ref{subsec:bg_decomp}.
We adapt it to treat each QA pair as an atomic sub-claim and compare against the decompositions of FactScore \citep{min-etal-2023-factscore} and R-ND (DecompScore’s own GPT-based method).

\paragraph{Setup.}
We randomly sample 1,000 sentences from the FactScore benchmark, following the evaluation protocol of \citet{wanner-etal-2024-decompscore} for comparability.
For each sentence, we generate question–answer pairs using our fine-tuned \texttt{LLaMA-3-8B} QA-Noun parser and the QA-SRL parser \citep{klein-etal-2022-qasem}.
Each QA pair is treated as a candidate meaning unit and evaluated for total count, non-redundant count after CORE filtering, and source entailment via DecompScore’s GPT-based pipeline.

Because the verbal and nominal parsers target different parts of speech, they occasionally recover overlapping semantic relations.
To avoid double-counting the same sub-claim, we use the \textbf{CORE} framework \citep{jiang-etal-2024-core} to automatically identify and cluster redundant or paraphrastic QA pairs.
This ensures a faithful count of unique atomic facts while still allowing cross-validation of equivalent relations across syntactic forms.
Notably, these overlaps reflect distinct yet complementary linguistic realizations of the same content, and may offer added value in enriching semantic labels for certain  downstream tasks.
See Appendix~\ref{app:overlap-examples} for examples and discussion of common overlap patterns.

Notably, unlike FactScore and R-ND GPT-4o–based baselines, our entire pipeline uses open-weight models to enable reproducible, scalable deployment.

\paragraph{Results.}

\autoref{tab:decomp_score} summarizes the results.
The combined QA-Noun+QA-SRL system yields over \textbf{150\%} more validated, non-redundant semantic units than FactScore and about \textbf{130\%} more than DecompScore, highlighting the granularity advantage of a structured predicate–argument approach over unconstrained fact extraction.
Nominal QA pairs account for nearly half of the decomposition output (2.4 QA pairs per sentence on average vs.\ 2.3 from verbs), underscoring the necessity of modeling noun-centered semantics for complete sentence-level meaning and cross-text alignment.

Although GPT-4o baselines achieve slightly higher entailment precision, our approach delivers substantially greater coverage and atomicity.
This reflects an inherent trade-off: exhaustively recovering predicate–argument structures for both verbs and nouns is a harder task than producing unconstrained facts, yet it yields decompositions that are more linguistically grounded and robust.
Importantly, many of the redundant QAs filtered by CORE are \textit{valid paraphrases} rather than noise, providing complementary surface realizations of the same fact --- an aspect future systems could exploit for richer semantic labeling or multi-view learning.

An illustration of the resulting decomposition is shown in \autoref{fig:decomp_example}.
Taken together, these findings position QA-Noun+QA-SRL --- and by extension the broader QASem framework --- as a comprehensive and structured alternative to fact-based decomposition methods for applications requiring precise cross-text semantic alignment.

\section{Conclusion}
\label{sec:conclusions}

We introduced QA-Noun, a framework for representing noun semantics as a set of QAs, each expressing a predicate-argument level fact involving the noun, which integrates seamlessly into the broader QA-based semantic paradigm.
By combining templated and open-ended questions, QA-Noun provides a scalable and interpretable method for capturing noun-centered meaning in context.
Our dataset and evaluations demonstrate its reliability and ability to capture rich, fine-grained relations, making it a strong foundation for highly-granular decomposition of textual meaning, as needed for cross-text semantic alignment.
Together with QA-SRL and related QA-based tasks, QA-Noun advances toward a comprehensive framework for sentence-level semantic decomposition.

\section*{Limitations}
\label{sec:limitations}
While QA-Noun provides a structured and scalable approach for semantic role labeling of noun predicates, several limitations remain in its current form.

\paragraph{Training Data Quality.}  
As discussed in Section~\ref{sec:dataset_construction}, our training data was single-annotated to enable greater diversity and scale within the available budget, while double annotation and adjudication were reserved for the evaluation sets.  
This practical design choice follows established practice but inevitably introduces some annotation noise and variability, particularly for subtle or context-dependent roles.  
Such inconsistencies may limit the ultimate performance of models trained solely on the training data, motivating future work on selective re-annotation or semi-automatic verification of difficult cases.

\paragraph{Question Template Design.}  
The fixed set of question templates enables consistency and interpretability, but also imposes important limitations. First, many of the templates are not phrased in fully natural language, which may hinder large language models (LLMs) that rely on surface form likelihoods for generation. This can lead to errors when models favor more common but less semantically appropriate templates. Second, the templates are not strictly mutually exclusive --- different questions can validly apply to the same argument span --- posing challenges during both training and evaluation. Disambiguating between overlapping templates remains a non-trivial problem for both humans and models.

\paragraph{Model Efficiency and Scale.}  
Our best-performing model in terms of raw F1 is a large in-context LLM (LLaMA 405B), which is expensive to run and unsuitable for deployment at scale. While our fine-tuned models (e.g., LLaMA 8B) offer a more practical solution, they still fall short of state-of-the-art LLMs in some scenarios.

\paragraph{Domain Generalization.}  
QA-Noun is built from and evaluated on formal text from Wikipedia and Wikinews. Although diverse, these sources do not capture the full spectrum of language use. It is unclear how well models trained on QA-Noun would generalize to other genres such as narrative, conversational, or low-resource domains.


\bibliography{custom}

\appendix

\section{Appendix}
\label{sec:appendix}
\begin{figure*}[t!]
    \centering
    \includegraphics[width=1\textwidth]{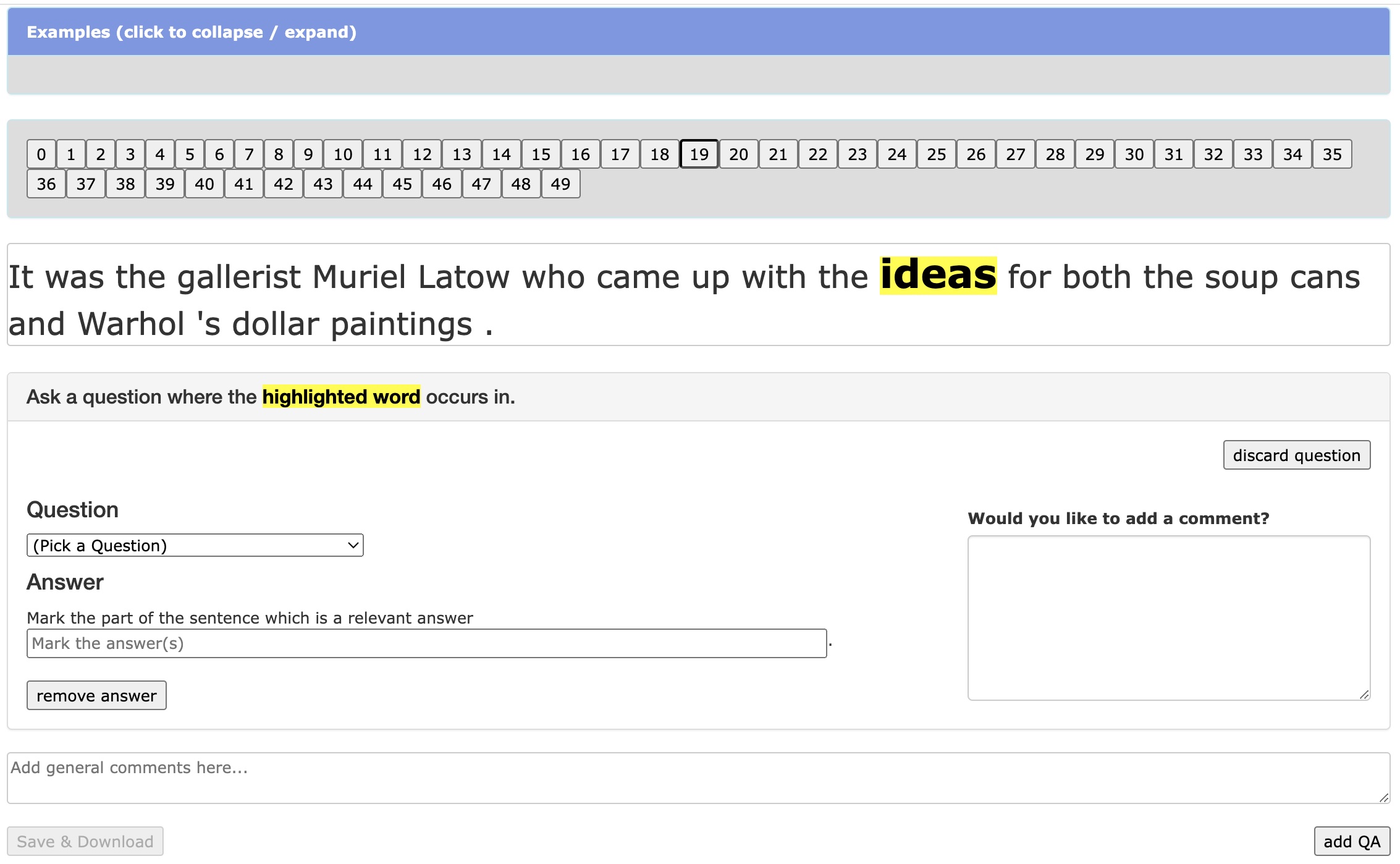}
    \caption{Graphical User Interface (GUI) for annotators.}
    \label{fig:gui_screenshot}
\end{figure*}
\begin{table*}[t!]
\centering
\begin{tabular}{p{0.96\textwidth}}
\hline
\scriptsize
\begin{tabular}[c]{@{}l@{}}Read the Sentence and focus on the noun that is marked with \textless{}f\textgreater{}\textless{}/f\textgreater{}.\\ Find all the words or short phrases which provide information about the noun entity of the noun marked in \textless{}f\textgreater{}\textless{}/f\textgreater - they are called arguments.\\ The arguments should be a continuous span from the sentence, they should appear with the exact words and order as appeared in the sentence.\\ Use the arguments you found as answers, and generate questions to match those answers.\\ The questions should be taken from the list of templates:\\ 1: What is the \textless{}property\textgreater{} of (the) \textless{}f\textgreater{}noun\textless{}/f\textgreater{}?,\\ 2: Whose \textless{}f\textgreater{}noun\textless{}/f\textgreater{}?,\\ 3: Where is the \textless{}f\textgreater{}noun\textless{}/f\textgreater{}?,\\ 4: How much /How many \textless{}f\textgreater{}noun\textless{}/f\textgreater{}?,\\ 5: What is the \textless{}f\textgreater{}noun\textless{}/f\textgreater{} a part/member of?,\\ 6: What/Who is a part/member of \textless{}f\textgreater{}noun\textless{}/f\textgreater{}?,\\ 7: What/Who is (the) \textless{}f\textgreater{}noun\textless{}/f\textgreater{}?,\\ 8: What kind of \textless{}f\textgreater{}noun\textless{}/f\textgreater{}?,\\ 9: When is the \textless{}f\textgreater{}noun\textless{}/f\textgreater{}?\\ The number marks the template's number.\\ The \textless{}f\textgreater{}noun\textless{}/f\textgreater{} should be replaced with the noun that is marked with \textless{}f\textgreater{}\textless{}/f\textgreater{} in the sentence.\\ The \textless{}property\textgreater tag should be replaced with a word that describes a property of the noun (color, size, cause etc.), that matches the answer.\\ Don't generate the same answer for two different questions, choose the most suitable question for each answer.\\ Display the list of QAs sorted in ascending order by question template id.\\ If you can't find any arguments to the noun marked in \textless{}f\textgreater{}\textless{}/f\textgreater{}, the output should be: "There are no QAs generated."\\ \\ The format should be:\\ QAs:\\ Question template number: \textless{}the number\textgreater\\ Question: \textless{}the question\textgreater\\ Answer: \textless{}the answer\textgreater{}\end{tabular} \\ \hline
\caption{Prompt used for in-context and fine tuning tasks}
\label{tab:in_context_prompt}
\end{tabular}
\end{table*}

\subsection{Model}
\label{sec:appendix-model}
The prompt used for the in-context and fine-tuning experiments is shown in \autoref{tab:in_context_prompt}.

\subsection{Fine-Tuning Configurations}
\label{sec:appendix-ft}

We performed a targeted hyperparameter search on the LLaMA 8B (v3) model, experimenting with a range of LoRA settings that varied in rank, scaling factor ($\alpha$), and training epochs (see \autoref{tab:ft_configs}). All experiments used a learning rate of 0.0002 and the AdamW optimizer.
For the larger models, Qwen 14B (v2.5) and Phi 14B (v4), we did not perform a full hyperparameter sweep due to compute constraints. Instead, we applied promising configurations observed during LLaMA tuning. For Qwen, we reused the best-performing LLaMA settings (rank 64, $\alpha$ 16), while for Phi we adopted a configuration (rank 32, $\alpha$ 8) that was reported to perform well in public benchmarks.

Among all configurations, the strongest results on the QA-Noun development set were achieved by:

\texttt{LlaMA-3-8B} with rank 64, $\alpha = 16$

\texttt{LlaMA-3-8B} with rank 8, $\alpha = 32$

\texttt{Phi-4-14B} with rank 32, $\alpha = 8$

The best-performing configuration on the QA-Noun development set was \texttt{LlaMA-3-8B} with rank 64 and $\alpha = 16$, which we selected for our parser.

\begin{table}[h]
\centering
\begin{tabular}{lccc}
\toprule
\textbf{Model} & \textbf{LoRA Rank} & \textbf{$\alpha$} & \textbf{Epochs} \\
\midrule
LLaMA 8B (v3)   & 64  & 16  & 6  \\
LLaMA 8B (v3)   & 32  & 128 & 3  \\
LLaMA 8B (v3)   & 64  & 64  & 3  \\
LLaMA 8B (v3)   & 16  & 64  & 10 \\
LLaMA 8B (v3)   & 64  & 16  & 3  \\
LLaMA 8B (v3)   & 16  & 32  & 20 \\
LLaMA 8B (v3)   & 8   & 32  & 3  \\
Qwen 14B (v2.5) & 64  & 16  & 3  \\
Phi 14B (v4)    & 32  & 8   & 3  \\
\bottomrule
\end{tabular}
\caption{LoRA fine-tuning hyperparameter configurations explored for each model.}
\label{tab:ft_configs}
\end{table}


\subsection{Annotation Interface}
\label{sec:appendix-annotation-interface}
We developed a dedicated Graphical User Interface (GUI) (see \autoref{fig:gui_screenshot}) that presented annotators with a list of sentences, each containing a highlighted noun, and tasked them with generating question-answer pairs specific to the noun.
To create the questions, annotators first selected an appropriate question template from a drop-down menu and populated any required slots for the selected template. 
Then, they marked the corresponding argument-answer by selecting a contiguous span of text within the sentence.






\subsection{Annotation Costs}
\label{sec:appendix-annotation-costs}
Our annotators were paid \$13 per hour for both generation and reconciliation steps, which is approximately 170\% of the local minimum wage. 
This resulted in an average cost of \$1.75 per predicate in the evaluation set --- compensating for two annotation steps and a reconciliation session.
For the training set, which employed a single annotator step, the average cost per predicate was \$0.35.
In total, the cost of dataset curation and annotator onboarding is estimated at approximately \$2,980.  


\begin{table*}[t!]
\small
\resizebox{\textwidth}{!}{%
\begin{tabular}{@{}p{0.6\textwidth}p{0.2\textwidth}@{}}
\toprule
\textbf{Sentence and QA Pairs} & \textbf{Category} \\
\midrule
\textbf{Sentence:} She has written articles and essays for various journals, edited volumes, and exhibition catalogs. \newline
QA-Noun: Whose \textbf{articles}? → \textit{She} \newline
QA-SRL: Who has \textbf{written} something? → \textit{She} & Agent \\
\midrule
\textbf{Sentence:} Father Tompkins also played a significant role in shaping the labor movement in Nova Scotia. \newline
QA-Noun: Whose \textbf{role}? → \textit{Father Tompkins} \newline
QA-SRL: Who \textbf{played} something? → \textit{Father Tompkins} \newline
QA-Noun: Where is the \textbf{movement}? → \textit{Nova Scotia} \newline
QA-SRL: Where did someone \textbf{play} something? → \textit{Nova Scotia} & Agent, Location \\
\midrule
\textbf{Sentence:} She served as Chair of the Department of Performing and Fine Arts from 2012 to 2019. \newline
QA-Noun: When is the \textbf{position}? → \textit{2012–2019} \newline
QA-SRL: When did someone \textbf{serve} as something? → \textit{2012–2019} & Time \\
\midrule
\textbf{Sentence:} Over the course of his career, Nieves played for several MLB teams including the New York Yankees and others. \newline
QA-Noun: Whose \textbf{teams}? → \textit{MLB (Nieves)} \newline
QA-SRL: What included something? → \textit{several MLB teams} & Possession \\
\midrule
\textbf{Sentence:} Tugman has utilized his skills to secure multiple victories on the professional circuit. \newline
QA-Noun: What is the \textbf{purpose} of the skills? → \textit{to secure victories} \newline
QA-SRL: Why has someone \textbf{utilized} something? → \textit{to secure victories} & Purpose \\
\bottomrule
\end{tabular}%
}
\caption{
Examples of overlapping semantic relations recovered by both QA-SRL (verbal) and QA-Noun (nominal) parsers and filtered by the CORE framework. Categories include Agent, Location, Time, and others such as Purpose and Possession.
}
\label{tab:core_redundant}
\end{table*}

\subsection{Overlap Between QA-Noun and QA-SRL}
\label{app:overlap-examples}

Although QA-Noun and QA-SRL annotate complementary syntactic structures, they often recover semantically equivalent relations.
To ensure accurate counts of unique content units during evaluation, we apply the \textbf{CORE} framework to detect and remove such redundancies.

In \autoref{tab:core_redundant} we present a range of overlap examples filtered out by CORE and categorize them into common types of semantic overlap.
We find that \textbf{Agent} overlaps --- e.g., a noun’s possessor versus the subject of a verb --- are the most common, reflecting shared underlying predicate-argument structures.
\textbf{Location} overlaps are also frequent when both noun phrases and verb predicates reference spatial context.
\textbf{Time} overlaps arise when temporal markers are linked to both event-denoting nouns and corresponding verbal mentions.
Other overlap types include \textbf{Purpose}, \textbf{Membership}, and \textbf{Possession}.
In all cases, CORE helps ensure that overlapping facts are scored only once while preserving their alignment for downstream applications such as paraphrase learning or redundancy detection.


\end{document}